\documentclass{Interspeech}

\interspeechcameraready 

\title{Swedish Whispers; Leveraging a Massive Speech Corpus for Swedish Speech Recognition}

\author{Leonora}{Vesterbacka}
\author{Faton}{Rekathati}
\author{Robin}{Kurtz}
\author{Justyna}{Sikora}
\author{Agnes}{Toftgård}
\affiliation{KBLab}{National Library of Sweden}{Sweden}
\email{\{leonora.vesterbackaolsson,faton.rekathati,robin.kurtz,justyna.sikora,agnes.toftgard\} @kb.se}

\usepackage{comment}
\usepackage{arydshln}
\usepackage{makecell}
\usepackage{multirow}
\begin{document}

\maketitle

\begin{abstract}
This work presents a suite of fine-tuned Whisper models for Swedish, trained on a dataset of unprecedented size and variability for this mid-resourced language.
As languages of smaller sizes are often underrepresented in multilingual training datasets, substantial improvements in performance can be achieved by fine-tuning existing multilingual models, as shown in this work.
This work reports an overall improvement across model sizes compared to OpenAI's Whisper evaluated on Swedish.
Most notably, we report an average 47\% reduction in WER comparing our best performing model to OpenAI's \texttt{whisper-large-v3}, in evaluations across FLEURS, Common Voice, and NST.
\end{abstract}

\section{Introduction}
The abundance of openly available audio data in English enables pre-training and fine-tuning of speech-to-text models on hundreds of thousands to millions of hours of recorded speech.
As a result, automatic speech recognition (ASR) systems are approaching human level robustness in English.
Other languages' performance in multilingual speech recognition tends to stand in proportion to the amount of data included from the language -- or the language family -- in question.
For low- to mid-resource languages with fewer speakers, such as Swedish, the amount of openly available data may be limited, and as a consequence these languages tend to be underrepresented in large-scale efforts to train multilingual speech recognition systems.
This results in suboptimal performance of ASR applications, such as voice-based customer service, and automatic captioning.
Language models and ASR models are often trained on massively multilingual datasets.
By continuing training these multilingual models for a specific language, the performance is greatly improved compared to the performance of multilingual models.
This is necessary for low- and mid-resource languages to remain competitive and sovereign in the rapid AI development of today.
Similarly to previous work for Norwegian~\cite{kummervold2024whispering}, we fine-tune Whisper for all canonical sizes on our massive dataset for Swedish.

Large collections of audio recordings maintained by cultural heritage institutions, such as the National Library of Sweden, and government agencies can potentially bridge some of the existing speech recognition performance gaps between Swedish and higher resourced languages.
By scaling up the amount of training data, one can noticeably improve Swedish speech recognition.
In this work, KBLab~\cite{CRL26325} at the National Library of Sweden have a have constructed a massive speech corpus with an emphasis on all variations of spoken Swedish, which is used to train speech recognition models for Swedish.

\section{Automatic speech recognition}
Automatic Speech Recognition models translate speech (i.e.\ spoken language) into text (i.e.\ written language).
The form of the output may vary between phonetic transcriptions, raw text without punctuation, and formatted sentences with punctuation.

\subsection{Model architecture}
Whisper~\cite{radford2023robust} is a multilingual ASR model with additional translation capabilities.
By relaxing the quality standard of the labeled training data, vast amounts of training data are unlocked.
The authors show that even with imperfectly labeled training data, such as subtitles, it is still possible to achieve speech recognition performance approaching human-level robustness in English.
Using the latest official Whisper models as a starting point, we fine-tune them on our herewith presented corpus for Swedish speech.

The Whisper model architecture is an encoder-decoder transformer~\cite{vaswani2017attention}.
The input audio is split up into \SI{30}{\second} chunks, resampled to \SI{16}{\kilo\hertz}, and transformed to an 80-channel log-magnitude Mel spectrogram (128-channel for \texttt{large-v3}).
Feature normalization is applied along with a two-layered 1-dimensional convolutional encoder with a GELU activation function, followed by sinusoidal position embeddings, and transformer encoder blocks.
We employ a Byte Pair Encoding (BPE) dropout~\cite{provilkov-etal-2020-bpe} of 0.2, which was shown to have beneficial regularizing effects for the large model size~\cite{kummervold2024whispering}.
An activation dropout of 0.1 is also applied.

The model is trained using the PyTorch implementation by Hugging Face that is detailed in the \emph{Whisper fine-tuning Event}\footnote{\url{https://github.com/huggingface/community-events/tree/main/whisper-fine-tuning-event}}.
We adapt this script and add functionality to train with timestamps as well as with previous contexts as a prompt.

\section{A massive Swedish speech corpus}
OpenAI's \texttt{whisper-large-v3}\footnote{\url{https://huggingface.co/openai/whisper-large-v3}} has been trained on \SI{680000}{\hour} of multilingual speech, of which \SI{2119}{\hour} are in Swedish.
While~\cite{radford2023robust} report that all increases in dataset size lead to improved performance, the improvement slightly stagnates after a dataset size of around \SI{50000}{\hour}.
With these numbers in mind, a dataset consisting of \SI{50000}{\hour} of transcribed Swedish was constructed.

Not only is the amount of data important to training ASR models, but also language variation.
One common drawback of using the multilingual Whisper is its inability to recognize variations in a lower resourced language, such as dialects.
While dialects are generally more difficult to correctly transcribe, given their decreased occurrence in the training data, this is especially noticeable for low- and mid-resource languages.
Models trained on material from the web, which can be assumed for models such as Whisper, have only learnt to recognize speakers that are commonly represented in web material, e.g.\ YouTube videos.
Training on this type of material excludes many varieties of speech that exists in a population. 
By actively including data sources that represent other types of speech, we increase the probability of developing a model that performs better for a wider variety of dialects, sociolects, age groups, and accents spoken by non-native Swedish speakers.

In this work, KBLab at the National Library of Sweden has collaborated with the Riksdag administration (i.e.\ the parliament administration in Sweden), Sweden's Television (SVT), and the Institute for Language and Folklore (Institutet för språk och folkminnen), and together we have been able to construct a transcribed Swedish dataset of unprecedented size and variability.
These dataset types can be split into four categories, which are presented below.

\subsection{Subtitles}
Subtitles have traditionally existed for Swedish television to subtitle non-Swedish TV broadcasts.
Swedish closed captions for Swedish spoken content have become increasingly available as a means to aid accessibility for the hearing impaired.
The Swedish subtitling tradition tends to condense the content rather than provide verbatim transcriptions, in order to provide the viewer enough time to read. 

In our filtering, we select only Swedish TV broadcasts and web content where metadata indicates the subtitle track is in Swedish. 
To ensure the speech is also in Swedish, we perform language detection using OpenAI's \texttt{whisper-large-v3}.
The final subtitle dataset is thus constructed out of audio identified as Swedish paired with Swedish subtitles. 

\subsubsection{SMDB}
With access to the Swedish Media Database (SMDB) at the National Library of Sweden, we extracted audio and subtitles from 15 Swedish TV channels, in a \SI{24}{\hour} stream format, without any program metadata, from 2021 to 2023, resulting in more than \SI{150000}{\hour} of raw broadcasts.
Of the \SI{63000}{\hour} of audio matching the subtitles, around 80\% were duplicates.
From this deduplicated subset, 20\% did not contain Swedish speech.
Finally, another 30\% of chunks were dropped, not reaching the desired quality for training.

\subsubsection{SVT}
In contrast to the SMDB dataset, cleaning the data provided by SVT was kept to a minimum, as it was delivered separated into various programs with reliable metadata.
Starting with \SI{30275}{\hour} of chunks, 61\% was kept for training.

\subsubsection{YouTube}
We additionally collected audio and subtitles from $2704$ YouTube channels resulting in about \SI{15416}{\hour}, before further quality filtering is applied using our pipeline described in Section~\ref{sec:datapreproc}.
Since subtitles on YouTube do not necessarily follow the same guidelines as TV, they generally provide a more verbatim transcription of what is spoken.

\subsection{Parliamentary recordings}
As a part of this project, we have been granted access to all digitized audio recordings of the Swedish Riksdag since 1966.
These recordings were matched and force-aligned against protocol transcripts of the debates, a part of the research corpus, the Swedish Parliament Corpus\cite{yrjanainen-etal-2024-swedish}.
The resulting speech corpus is named \textsc{RixVox-v2}, and consists of \SI{18000}{\hour} of transcribed speech recorded before 2003 and \SI{5500}{\hour} after 2003.
Apart from its sheer size, this corpus also exhibits a wide variation in Swedish dialects as well as providing a both modern and historic resource for political spoken Swedish.
The total amount of audio recordings used in this project is \SI{23500}{\hour}, of which almost \SI{22000}{\hour} pass the ``Stage 1'' quality filters outlined in Section~\ref{sec:datapreproc}.
We publish the entire dataset \textsc{RixVox-v2} on Hugging Face with a permissive license to further research efforts in Swedish speech technology.

\subsection{Dialect recordings}
To improve ASR performance on Swedish dialects, special dialect recordings were added to the training data.
The Institute for Language and Folklore is a Swedish government agency that hosts a large collection of speech recordings.
Some of these recordings are the results of research projects, such as SweDia~\cite{5380888}, in which various Swedish dialects have been professionally recorded and transcribed.
Although this corpus does not compare in size with the other datasets, they provide an important diversity in its speech, needed for developing a model which is representative of the whole population.
\subsection{Other}

\subsubsection{NST}
The NST dataset, collected by Nordic Language Technology (Nordisk Språkteknologi), is hosted by Language Bank at the National Library of Norway\footnote{\url{https://www.nb.no/sprakbanken/en/resource-catalogue/oai-nb-no-sbr-56/}}.
It consists of hundreds of hours of recordings and transcriptions of up to 1000 speakers with different Swedish dialects.
The recordings consist of both longer sentences as well as single-word examples.
The dataset is split into a training set which is included in our training data and a test set which is used as a hold-out test set in our evaluation.

\subsubsection{Common Voice}
Mozilla is collecting an open source, crowd-sourced, multilingual ``dataset of voices'' called \emph{Common Voice}~\cite{ardila2020common}.
The dataset is the largest publicly available speech dataset, of which Swedish constitutes \SI{54}{\hour} of recordings.
The train, validation, and test splits of Common Voice 16.1 are all used as hold-out test sets.

\subsubsection{FLEURS}
FLEURS~\cite{conneau2022fleurs}, \emph{Few-shot Learning Evaluation of Universal Representations of Speech}, is an n-way parallel speech dataset in 102 languages, with approximately \SI{12}{\hour} of speech per language.
We use the FLEURS validation split for validation purposes during training, but keep its train and test splits as hold-out test sets for our final evaluation.

\subsection{Data preprocessing}\label{sec:datapreproc}

The preprocessing pipeline is common to all dataset types and is outlined below:
\begin{enumerate}
\item The audio is kept in its source format and converted to \SI{16}{\kilo\hertz} mono slices on the fly during all processing steps.
A corresponding format-agnostic metadata file for each audio file contains the raw units of observation: e.g.\ subtitle blocks or force aligned sentences, using a single unified format.
\item Only Swedish text transcriptions are retained.
We filter based on subtitle track metadata for data sources whose metadata is reliable.
Language detection classifiers for text are used where metadata is not reliable.
\item Candidate chunks to be used during training are created by combining and packing our observations to chunks of up to \SI{30}{\second} length based on timestamp metadata. 
Shorter chunks are additionally sampled by setting a smaller maximum chunk length, in order to boost the model's robustness to variable audio input lengths.
\item Audio language detection is performed on all candidate chunks using \texttt{whisper-large-v3}. Only chunks where Swedish is detected are retained. 
\item Candidate chunks passing the previous filter are transcribed using \texttt{whisper-large-v3} and \textsc{VoxRex}, a Swedish Wav2Vec2.0\cite{malmsten2022hearingvoicesnationallibrary}.
\item The transcribed results are compared to the original transcript using the following metrics:
\begin{itemize}
\item The \emph{Word Error Rate} (WER) is a common metric derived from the Levenshtein distance.
It measures how accurately an ASR model transcribes spoken language compared to a reference transcript.
A lower WER corresponds to a better transcription compared to the reference text.
The Character Error Rate (CER) is the character-level equivalent of WER.
\item The \emph{Bilingual Evaluation Understudy} (BLEU) score~\cite{papineni-etal-2002-bleu}, is a precision-based metric that computes a value between 0 and 1, where 1 corresponds to a perfect agreement between the prediction and the gold standard.
\item \emph{Recall-Oriented Understudy for Gisting Evaluation} score (ROUGE-N)~\cite{lin2004rouge} is a recall-based metric that measures the overlap of n-grams between a model's output and a reference text.
\item  Approximate string matching is the technique of finding strings that match a pattern approximately (rather than exactly). This is used to match the first and last words in a chunk. We additionally calculate a CER based metric to match the first and last 10 characters in a chunk. 
\end{itemize}
\end{enumerate}

Using the above metrics to measure the quality of the transcription, we create two categories for our training corpus.
Data that do not meet the quality filters applied in either category are excluded from the training.

\textbf{Stage 1}
The first category is defined by a set of relaxed values for CER and BLEU, comparing the original transcription to that of \texttt{whisper-large-v3} and \textsc{VoxRex}.
Although the \texttt{whisper-large-v3} transcriptions are generally of higher quality than those from \textsc{VoxRex}, both transcriptions are used to safeguard against potential hallucinations from Whisper.
We apply dataset specific BLUE thresholds based on qualitative assessments, where the main aim during ``Stage 1'' is to filter out most of the low signal and pure noise observations where ground truth has very low or no overlap with the audio.
CER thresholds comparing the first and last 10 characters of ground truth and machine transcriptions are used only to determine which samples to train with timestamp tokens during this stage.   
The hypothesis behind this category of data is that even though the transcriptions are not perfect, they are still good enough to learn from. 

\textbf{Stage 2}
The second category, denoted ``Stage 2'', is characterized by a set of higher requirements for the quality of the transcriptions, to ensure that only the most accurate examples from the training data are included in the second stage.
If the previous category was characterized by the width of the variation of Swedish, this category will instead be characterized by the transcription style.
Additionally, in Stage 2 filtering, we introduce another metric, ROUGE-N, to further ensure high transcription accuracy.
During filtering, we calculate BLEU score and a weighted combined ROUGE-N score using weights that place higher emphasis on longer n-grams.
We assign weights 0, 0.25, 0.5, 0.25, respectively to 1-, 2-, 3- and 4-grams.
A CER threshold of 0.2 or lower for the first and last 10 characters is used here to determine whether to include the sample. 

The final composition of our corpus is shown in Table~\ref{tab:datacomposition}.

\begin{table}[th]
  \caption{Composition of our Swedish speech corpus}
  \label{tab:datacomposition}
  \centering
  \begin{tabular}{l l l l}
    \toprule
    \textbf{Dataset} & & \textbf{Stage 1 (h)} &\textbf{Stage 2 (h)} \\\hline
     \multirow{3}{*}{Subtitles} & SVT          & $18\,468$   &  $688$    \\
     & SMDB         & $8\,629$   &  $421$       \\
     & YouTube      & $7\,164$     &  $2\,001$        \\
           \\
    Riksdag  &      & $21\,949$ & $5\,119$ \\
     Dialects      & & 54     & 54 \\  
    NST & & 250    & 250 \\\hline
    Total  &         & $56\,514$ & $8\,533$ \\\bottomrule
  \end{tabular}
\end{table}
\subsubsection{Non-speech and timestamps}
Although the above pipeline would filter out any examples where there is no speech, we actively include some portion of the data where no speech is present.
One common issue with Whisper is its hallucinations, where the model outputs transcriptions not corresponding to the speech, or transcriptions when there is no speech.
One hypothesis behind hallucinations is the lack of training examples with no speech and no corresponding transcriptions.
To this end we include examples in our training data with no speech or transcription to help the model learn not to output any transcriptions when there is silence.
The total amount of non-speech in our training data is \SI{100}{\hour}.

A portion of the training data is prepared with timestamp tokens added at the subtitle block or sentence level, with the goal of retaining the original Whisper model's ability to output timestamps. 
In order to provide a consistent signal to the model, we only train with timestamps on those examples where the CER between ground truth and the first and last 10 characters of an ASR transcription is equal to 0.2 or lower.
Timestamps are sampled into the training at a rate of roughly 50\% during ``Stage 2'', and at lower single digit percentage rates in ``Stage 1''.

\subsubsection{Previous context as prompt}

The datasets are processed to allow for the option to train on the text of a previous adjoining audio segment as a contextual prompt. We train with the previous context at a 50\% rate. 
Since Whisper uses the same token, \texttt{<|endoftranscript|>}, for padding and end of transcript, a crucial detail in loss masking implementation is to leave one \texttt{<|endoftranscript|>} unmasked, while masking the last $n-1$ occurences.

\section{Model training}
In~\cite{radford2023robust} five model sizes of Whisper were trained.
In order to accommodate all end users, from large scale transcription tasks utilizing GPUs to modest applications run on a laptop, all model sizes are fine-tuned.
The model sizes in parameters range from 39 million to 1.5 billion and are outlined in Table~\ref{tab:modelsizes} along with the chosen learning rates for the training.
\begin{table}[th]
  \caption{Number of parameters, learning rate and node hour consumption for all Whisper model sizes}
  \label{tab:modelsizes}
  \centering
  \begin{tabular}{ l l  l l}
    \toprule
    \textbf{Model name} & \textbf{Parameters} & \textbf{Learning rate} & \textbf{Node hours} \\
    \midrule
  tiny & 39 M & $6 \times 10^{-4}$  &  600 \\
 base  & 74 M     & $4 \times 10^{-4}$  &   700   \\
  small & 244 M   &$2 \times 10^{-4}$  & 880\\
  medium  & 769 M    &$5 \times 10^{-5}$ &  1600      \\
 large  & 1.55 B   &$7 \times 10^{-5}$  &  3100       \\
    \bottomrule
  \end{tabular}
\end{table}
Much of the training details are similar to those detailed in~\cite{radford2023robust}, such as AdamW~\cite{loshchilov2018decoupled} optimizer and a linearly decaying learning rate after 5000 steps.
The main modifications with respect to the original Whisper training lay in the choice of learning rate, that we set to slightly lower values. 
All model sizes are trained with a two-stage approach, an initial training using the Stage 1 portion of the data for $150\,000$ steps, with the corresponding learning rates detailed in Table~\ref{tab:modelsizes}.
This initial training stage is followed by a second training stage for 50 000 steps using the Stage 2 category of the data.

All models are trained on the \textsc{Leonardo} supercomputer~\cite{turisini2023leonardo}, with access awarded through the EuroHPC Joint Undertaking.
Each computing node at the \textsc{Leonardo} supercomputer is equipped with four Nvidia custom A100 64GB GPUs.
The total computing consumption is detailed in Table~\ref{tab:modelsizes}.
The total train batch size is set to 1024, with per-device batch sizes of 16.

\section{Results and discussion}
We report test results as a comparison between OpenAI's Whisper and our models evaluated using the metrics word error rate (WER) and BLEU scores.
The splits of three hold-out datasets which have not been employed for validation purposes during training are used: namely FLEURS (train and test set), NST (test set), and Common Voice (train, validation, and test set).
Common Voice and FLEURS are not part of our training data and as such they represent the models' out-of-domain performance.
There is a significant improvement in performance for all model sizes trained in this paper compared to OpenAI's Whisper, which can be seen in Table~\ref{tab:WER} for WER and Table~\ref{tab:BLEU} for BLEU.

\begin{table}[th]
  \caption{Model performance measured in WER (lower is better) evaluated on FLEURS, CommonVoice, and NST.}
  \label{tab:WER}
  \centering
  \begin{tabular}{l l l l l}
    \toprule
    \textbf{Model size } & & \textbf{FLEURS} & \textbf{\makecell{Common-\\Voice}} & \textbf{NST}\\\hline
    
    \multirow{2}*{tiny} & KBLab & \textbf{13.2}  & \textbf{12.9} & \textbf{11.2}  \\ 
     & OpenAI & 59.2 & 67.8 & 85.2 \\\hline
    
    \multirow{2}*{base} &KBLab     & \textbf{9.1}  & \textbf{8.7} & \textbf{7.8}  \\ 
    & OpenAI & 39.6 & 52.1 & 53.4 \\\hline
    
    \multirow{2}*{small} & KBLab     & \textbf{7.3}  & \textbf{6.4} & \textbf{6.6}  \\ 
    & OpenAI & 20.6 & 26.4 & 26.4 \\\hline
    
    \multirow{2}*{medium} & KBLab      & \textbf{6.6}  & \textbf{5.4} & \textbf{5.8}  \\ 
    & OpenAI & 12.1 & 15.8 & 17.1 \\\hline
    
    \multirow{2}*{large-v3} & KBLab     & \textbf{5.4}  & \textbf{4.1} & \textbf{5.2}  \\ 
    & OpenAI & 7.8 & 9.5 & 11.3 \\
    
    \bottomrule
  \end{tabular}
\end{table}

\begin{table}[th]
  \caption{Model performance measured in BLEU (higher is better) evaluated on FLEURS, CommonVoice, and NST.}
  \label{tab:BLEU}
  \centering
  \begin{tabular}{l l l l l}
    \toprule
    \textbf{Model size } & & \textbf{FLEURS} & \textbf{\makecell{Common-\\Voice}} & \textbf{NST}\\\hline
    \multirow{2}*{tiny} & KBLab     & \textbf{76.6}  & \textbf{73.7} & \textbf{74.3}  \\ 
    & OpenAI & 26.9 & 21.1 & 24.0 \\\hline
    \multirow{2}*{base} & KBLab     & \textbf{83.2}  & \textbf{79.9} & \textbf{78.3}  \\ 
    & OpenAI & 41.1 & 32.5 & 36.9 \\\hline
    \multirow{2}*{small} & KBLab     & \textbf{86.6}  & \textbf{83.5} & \textbf{79.6}  \\ 
    & OpenAI & 64.0 & 56.5 & 58.2 \\\hline
    \multirow{2}*{medium} & KBLab     & \textbf{87.6}  & \textbf{85.0} & \textbf{80.2}  \\ 
    & OpenAI & 77.1 & 70.1 & 68.9 \\\hline
    \multirow{2}*{large-v3} & KBLab     & \textbf{89.8}  & \textbf{87.2} & \textbf{81.1}  \\ 
    & OpenAI & 84.9 & 79.1 & 75.1 \\
    \bottomrule
  \end{tabular}
\end{table}

Most notable is the improvement in WER for the smaller model sizes, where the WER for the OpenAI Whisper-tiny model evaluated on FLEURS resulted in 59.2, a number which has rendered the model more or less inutile.
We report an improvement in WER to 13.2 for the smallest model size (tiny).
The second noteworthy observation is that our Whisper-small outperforms OpenAI's Whisper-large as evaluated on these benchmarks.
This result has a positive implication on the amount of compute needed for transcription tasks, as the same performance can be achieved for a much smaller model that in turn consumes significantly less compute during inference. Most notably, we report an average 47\% reduction in WER comparing our best performing model to OpenAI's Whisper large-v3, in evaluations across FLEURS, Common Voice, and NST.

\subsection{Future work}
Although the results reported in this paper exhibit a massive improvement in ASR for Swedish, we see many avenues for future work that can further strengthen ASR for Swedish.

The data preprocessing pipeline as outlined in Section~\ref{sec:datapreproc} is relying on the transcriptions from OpenAI's \texttt{whisper-large-v3} to select high-quality training data.
An obvious drawback with this approach is that one can not differentiate if the training data is of bad quality (e.g.\ not identified as Swedish) or if the model used to transcribe is not good enough.
One risk is that this approach is rejecting training data with difficult dialects as the OpenAI model has not been trained on this type of speech before.
By using our fine-tuned Whisper-large for the transcription step in the data preprocessing pipeline, one could potentially select more high-quality training data as well as training data with rare dialects.
So data that had otherwise been filtered out due to a bad model performance, could in fact be included in the training data now, and further widen the range of quality in the training data.
With the same reasoning, the use of a better Wav2Vec2.0 model than the \textsc{VoxRex} used in the data preprocessing pipeline would be beneficial.

\section{Acknowledgements}

We acknowledge the EuroHPC Joint Undertaking for awarding this project access to the EuroHPC supercomputer LEONARDO, hosted by CINECA (Italy) and the LEONARDO consortium, through the Development Access call and AI and data intensive applications access call.

\bibliographystyle{IEEEtran}
\bibliography{mybib}

\end{document}